\documentclass[runningheads]{llncs}

\usepackage[T1]{fontenc}
\usepackage{graphicx}
\usepackage{tabularx}
\usepackage{multirow}
\usepackage{subcaption}
\usepackage{svg}
\usepackage{hyperref}
\usepackage{amsmath}
\usepackage{amssymb}
\usepackage{algorithm}
\usepackage{algpseudocode}

\begin{document}

\title{Distribution-Controlled Client Selection to Improve Federated Learning Strategies}
\titlerunning{Distribution-Controlled Client Selection to Improve FL}

\author{Christoph Düsing \and
Philipp Cimiano}
\authorrunning{C. Düsing \& P. Cimiano}

\institute{CITEC, Bielefeld University, Bielefeld, Germany \\ 
\email{\{cduesing, cimiano\}}@techfak.uni-bielefeld.de}

\maketitle

\begin{abstract}
Federated learning (FL) is a distributed learning paradigm that allows multiple clients to jointly train a shared model while maintaining data privacy. Despite its great potential for domains with strict data privacy requirements, the presence of data imbalance among clients is a thread to the success of FL, as it causes the performance of the shared model to decrease. To address this, various studies have proposed enhancements to existing FL strategies, particularly through client selection methods that mitigate the detrimental effects of data imbalance. In this paper, we propose an extension to existing FL strategies, which selects active clients that best align the current label distribution with one of two target distributions, namely a balanced distribution or the federations combined label distribution. Subsequently, we empirically verify the improvements through our distribution-controlled client selection on three common FL strategies and two datasets. Our results show that while aligning the label distribution with a balanced distribution yields the greatest improvements facing local imbalance, alignment with the federation's combined label distribution is superior for global imbalance. 

\end{abstract}

\section{Introduction}

In recent years, we have witnessed an increasing demand for data privacy and privacy-preserving machine learning \cite{truex2019hybrid}. In this regard, federated learning (FL) has emerged as a promising distributed learning paradigm that facilitates joint training of a shared model without compromising data privacy \cite{mcmahan2017communication}. 
Towards this end, FL relies on a central server to orchestrate the training procedure that consists of several rounds \cite{kairouz2021advances}. Instead of collecting all clients' data at the central server and thus exposing potentially sensitive and valuable data, FL initiates a global model, asks participating clients to fit them to their local data, and collects and aggregates the local model updates to a new global model afterwards \cite{kairouz2021advances,mcmahan2017communication}. 
Unfortunately, one of FL's major advantages, i.e., privacy preservation, comes at the cost of data quality assurance, exposing it to the threat of data imbalance among clients. Here, data imbalance refers to client-specific data properties that differ from the federation \cite{li2022federated}, which negatively affects the shared model's performance \cite{zhao2018federated,li2022federated,dusing2022trade}.
Therefore, multiple studies have proposed improvements to existing FL strategies to better cope with data imbalance (e.g., \cite{ou2022aggenhance,duan2019astraea,zhou2023federated,tsouparopoulosimproving}).

Here, client selection approaches, in particular, have demonstrated their suitability to mitigate its negative impact on performance \cite{wang2022federated,wang2023distribution}. Unlike previous approaches that randomly select the clients that participate in a round of federated training, client selection defines certain criteria to actively select the most promising clients for each round \cite{nishio2019client}. 
\textit{FedCor}, for example, selects clients with respect to their local loss to improve model convergence \cite{tang2022fedcor}. Other works rely on clients' data distribution as selection criteria, e.g., by introducing a label skewness measurement \cite{wang2022federated} or clients' pairwise data similarity \cite{wang2023distribution}.

Accordingly, client selection based on various measurements of data distributions has demonstrated promising results. However, to the best of our knowledge, no prior research considered aligning the overall label distribution during active client selection with pre-defined target distributions.
To this end, we propose a light-weight extension to existing FL strategies that selects clients such that they contribute to bringing the current label distribution closer to a pre-defined target distribution.
More precisely, the contributions of our work are as follows: 

\begin{enumerate}
    \item We introduces a greedy distribution-controlled (DC) client selection extension applicable to arbitrary FL strategies and demonstrate the improvements in terms of predictive performance in face of two different notions of data imbalance for the FL strategies \textit{FedAvg} \cite{mcmahan2017communication}, \textit{FedAtt} \cite{ji2019learning}, and \textit{FedProx} \cite{li2020federated};
    \item We propose, rigorously test, and discuss the use of two different target label distributions, namely \textit{Real} and \textit{Balanced} that control the client selection. Our findings prove that while the former one is more suitable to address global imbalance, the latter one is ideal for local imbalance;
    \item We conduct additional empirically analyses to verify the significance of our findings, identify the optimal number of clients to select, and compare our greedy selection to an exhaustive gold standard. 
\end{enumerate}


\section{Related Work} \label{related_work}

FL has emerged as a paradigm-shifting approach to distributed machine learning, emphasizing privacy and decentralization. By enabling multiple clients to collaboratively train a shared model while keeping their data localized \cite{mcmahan2017communication}, FL addresses the growing demand for data privacy \cite{chajewska2023federated}. FL's training procedure consists of several rounds of training. Each round involves clients receiving the shared model from the central server, fitting it to their respective local data, and sending only the model updates back to a central server \cite{mcmahan2017communication,kairouz2021advances}. The server then aggregates these updates to refine the global model, where the aggregation method itself is defined by the respective FL strategy in use \cite{mcmahan2017communication,ji2019learning,li2020federated}. 

In spite of its potential, FL faces significant challenges, particularly with data heterogeneity and imbalance across clients \cite{li2022federated}. Data imbalance refers to the situation where the distribution of data samples is uneven among clients, often leading to skewed model performance \cite{li2022federated,zhao2018federated}. This imbalance can occur in various forms, such as quantity imbalance, feature imbalance, or label imbalance \cite{dusing2022trade,wang2021addressing,buda2018systematic}. As a result, models trained federately under these conditions tend to perform well for some clients but poorly for those with atypical data distributions \cite{li2021survey,dusing2022trade}.
Hence, addressing data imbalance is crucial for the robustness and fairness of FL models. Several strategies have been proposed to mitigate this issue. These include data augmentation techniques \cite{ou2022aggenhance,duan2019astraea}, re-weighting of client contributions \cite{zhou2023federated}, and federated ensemble learning \cite{tsouparopoulosimproving}. However, these methods often come with trade-offs in terms of complexity \cite{deng2023enhancing}, computational overhead, or privacy risks \cite{guo2023fedbr,sattler2019robust}. Therefore, finding efficient and scalable solutions to data imbalance remains an active area of research in the FL community.

One promising approach to mitigating data imbalance in FL is client selection. Client selection techniques aim to actively choose a subset of clients for each training round, thereby ensuring a more balanced or otherwise beneficial participation \cite{nishio2019client}. By carefully selecting clients based on certain criteria -- such as data diversity, contribution to the global model, or resource availability -- these methods can enhance the model's generalization and fairness \cite{zou2021feddcs,wang2023distribution}.
One such approach leverages importance sampling, where clients with more significant or underrepresented data are given higher probabilities of being selected \cite{wang2020optimizing}. 
Other studies facilitate federated analytics \cite{wang2021federated} to calculate a skewness score for each client indicating their respective degree of imbalance \cite{wang2022federated}. Likewise, Wang et al. \cite{wang2023distribution} propose to leverage kernel-operations to project each client's feature distributions to a pairwise distribution score in order to reduce the discrepancy of data distributions during each round of training. 
More recently, Sheng et al. \cite{sheng2023modeling} facilitate a generative adversarial network to resemble the federation's combined data distribution, which allows to reduce performance degradation due to data imbalance.
Alternatively, clients can be clustered according to their label distributions. Each resulting cluster can afterwards federately train a shared model, leading to overall performance improvements \cite{diao2024exploiting}.
Finally, Ma et al. \cite{ma2021client} group clients according to their label quantity information and select clients such that their combined labels approximate the federation's global label distribution.

Despite the fact that existing client selection approaches demonstrate promising performance improvements, actively selecting clients to approximate a specific target label distribution as well as identifying the unique advantages of different target distributions remain open issues which we address in this work.

\section{Distribution-controlled Client Selection} \label{methodology}

In this work, we propose an extension to existing FL strategies to incorporate DC client selection at the start of each round of federated training. The rationale of the proposed approach and its three steps is the following: existing FL strategies such as \textit{FedAvg} \cite{mcmahan2017communication}, \textit{FedAtt} \cite{ji2019learning}, and \textit{FedProx} \cite{li2020federated} select a set of clients to actively participate in each round at random. In face of data imbalance and label imbalance in particular, this can lead to poor performance or convergence \cite{li2022federated}. Therefore, we propose to improve the overall label distribution of all active clients in each round. To do so, we either align their label distribution with a \textit{balanced} distribution (i.e., same number of samples for each class) or the \textit{real} federation's combined label distribution by adding favorable clients to the set of active clients. With that, we aim to smoothen the model training and thus improve the overall performance. Our proposed approach comprises three steps, namely the \textit{Target Distribution Selection}, the \textit{Client Selection}, and finally the \textit{Federated Training}. The procedure is outlined in more detail in Algorithm \ref{alg:dc}.
\begin{center}
\scalebox{0.75}{
{\centering
\begin{minipage}{0.8\linewidth}
\begin{algorithm}[H]
    \caption{DC client selection applied to an FL strategy}\label{alg:dc}
    \begin{algorithmic}[1]
    \Procedure{Server Execution}{}
        \State initialize $w_0$
        \State $\vec{V}_{target} \gets$ (select $\vec{V}_{Real}$ or $\vec{V}_{Balanced}$) \Comment{Step 1}
        
        \For{each round $t=1,2,...$}
            \State $S_t \gets$ \{random set of \textit{m} clients\}

            \For{$i = 1$ \textbf{to} $m_{DC}$ \textbf{step} $1$} \Comment{Step 2}
                \State $S_t \gets S_t + $ (get client from Equation \ref{eq:selection})
            \EndFor
            
            \For{each client $k \in S_t$ \textbf{in parallel}} \Comment{Step 3}
               \State $w_{t+1}^k \gets$ ClientUpdate$(k,w_t)$ 
            \EndFor
            \State $w_{t+1} \gets$ (FL strategy specific aggregation)
        \EndFor
    \EndProcedure
    \\
    \Function{Client Update}{$k,w$}
        \State $\mathcal{B} \gets$ (split $\mathcal{D}_k^{Train}$ into batches of size $B$)
        \For{each local epoch}
            \For{batch $b \in \mathcal{B}$}
                \State $w \gets w - \eta \nabla l(w,b)$
            \EndFor
        \EndFor
        \State return $w$ to server
    \EndFunction
    \end{algorithmic}
\end{algorithm}
\end{minipage}
\par
}
}
\end{center}
\textit{\textbf{Step 1: Target Distribution Selection.}} At first, we have to define the target distribution used to align the active clients' label distribution with. 
Therefore, we propose two different target distributions. The \textit{real} distribution reflects the federation's combined label distributions whereas the \textit{balanced} distribution expects all classes to be equally represented.

In order to compute the target distribution $\vec{V}_{real}$ without requiring clients to reveal their individual label distributions, which would violate their data privacy, we suggest to employ secure aggregation \cite{bonawitz2017practical,li2021secure}. This method allows to collect properties such as the label distribution from mutually distrustful clients as aggregates \cite{bonawitz2017practical}. To do so, it requires clients to mask their label distributions such that the central server can only access the aggregate of client-provided distributions \cite{li2021secure}. Accordingly, using secure aggregation, the \textit{real} label distribution is computed as follows: 
\begin{equation}
\vec{V}_{real}=[\sum_{c} N_{c}^{1},...,\sum_{c} N_{c}^{Q}], \label{eq:real}
\end{equation}
where $Q$ is the number of classes and $N_{c}^{q}$ the number of samples for class $q$ held by client $c$.
The target distribution $\vec{V}_{balanced}$ is significantly less complex to produce. Instead of relying on secure aggregation, it is defined as a vector of length $Q$ containing only \emph{1}s, i.e., $\vec{V}_{balanced}=[1,...,1]$. This simple yet effective approach ensures that all classes are expected to be equally represented in the set of active clients.

Finally, depending on the expected outcome and preferred choice of target distribution, $\vec{V}_{target}$ is either set to $\vec{V}_{real}$ or $\vec{V}_{balanced}$.
\\
\textit{\textbf{Step 2: Client Selection.}} The client selection is performed in each round of training and comprises of two sub-steps. At first, $m$ clients are randomly selected to actively participate in the round. We decided to keep this random selection and only subsequently start with the targeted client selection, as this random selection is a fundamental part in most FL strategies that we build upon. Furthermore, randomization ensures that all clients get to contribute to the federated training.
Subsequently, the client selection based on the previously selected target distribution $\vec{V}_{target}$ is performed. It follows a greedy procedure of iteratively adding one client to the set of active clients until a predefined maximum $m_{DC}$ is met or no more improvement in distribution alignment is achieved. 
Per iteration, the label distribution of all currently active clients $\mathbf{V}_{\text{active}}$ can be calculated using secure aggregation, similar to how it is done in step 1 and Equation \ref{eq:real}. 
Afterwards, $\mathbf{V}_{\text{active}}$ is propagated to all currently inactive clients. Then, each client $c \in \mathcal{C}$ first adds their respective local label distribution $\mathbf{V}_c$ to $\mathbf{V}_{\text{active}}$ before calculating its cosine distance to the target distribution $\mathbf{V}_{\text{target}}$. Note that the norm of the two candidate distributions $\vec{V}_{real}$ and $\vec{V}_{balanced}$ is significantly different but does not affect the client selection due to our choice of cosine distance as distance measure.
Afterwards, the central server determines the optimal client $c^{*}$ that provides the minimal cosine distance following Equation \ref{eq:selection}. Finally, $c^{*}$ is added to the set of active clients before the next iteration.
\begin{equation}
c^{*} = \arg \min_{c \in \mathcal{C}} \left( 1 - \frac{\left( \mathbf{V}_c + \mathbf{V}_{\text{active}} \right) \cdot \mathbf{V}_{\text{target}}}{\left\| \mathbf{V}_c + \mathbf{V}_{\text{active}} \right\| \left\| \mathbf{V}_{\text{target}} \right\|} \right) \label{eq:selection}
\end{equation}
\\
\textit{\textbf{Step 3: Federated Training.}} For the federated training and the aggregation of model updates specifically, we rely on three commonly used FL strategies. In the following, we briefly introduce each of them and elaborate on the integration of DC client selection to them.
\begin{itemize}
    \item \textbf{FedAvg.} Despite its known tendency to significantly suffer from data imbalance, FedAvg \cite{mcmahan2017communication} is the de-facto standard FL strategy due to its popularity and widespread application. It aggregates model updates using the mean weighted by the respective client's relative number of samples. 
    \item \textbf{FedAtt.} FedAtt \cite{ji2019learning} extends FedAvg by incorporating an attention mechanism that considers each client's contribution to the global model and thereby minimizes the distance between local models. It shows improved performance, especially on imbalanced data \cite{ji2019learning} 
    \item \textbf{FedProx.} Li et al. \cite{li2020federated} introduce FedProx as an extension to FedAvg that tackles the issue of data imbalance in FL. Therefore, it exhibits both more stable convergence and higher predictive performance.
\end{itemize}
In the following, we will evaluate the improvements through DC client selection for these three FL strategies. We will respectively refer to them as $FedAvg_{DC}$, $FedAtt_{DC}$, and $FedProx_{DC}$\footnote{A local implementation of $FedAvg_{DC}$ can be found \href{https://github.com/cduesing/FedAVGdc}{\underline{here}}.} when client selection is applied.

\section{Experiments} \label{eyxperiments}

\subsection{Experimental Setup}

In the following, we outline all relevant information on data and models to follow and reproduce the experiments we conducted. 
\\
\textit{\textbf{Datasets.}} For our study, we selected the two datasets \textit{MNIST} \cite{deng2012mnist} and \textit{CovType}\footnote{\label{note1}Received from the \href{https://archive.ics.uci.edu/datasets}{\underline{UCI Machine Learning Repository}}.}, both of which are publicly available and frequently used in FL research, especially when simulating data imbalance \cite{li2022federated,dusing2022towards,ou2022aggenhance}. 
MNIST contains about 70,000 images of handwritten digits (zero to nine) in a 28*28 resolution, with the same amount of samples for each of the ten classes. The dataset CovType contains 581,012 samples of tabular data, each of which has 54 features and belongs to one of two classes. Here, there is marginal imbalance between the classes, with about 52\% of samples belonging to the majority class.  
This choice of datasets requires binary as well as multi-class classification models on the one hand, and multi-layer perceptrons (MLPs) for tabular data as well as convolutional neural networks (CNNs) for image data on the other hand. 
\\
\textit{\textbf{Data Setting.}} We start by randomly splitting each dataset $\mathcal{D}$ into $\mathcal{D}_{Train}$ (80\%) and $\mathcal{D}_{Test}$ (20\%). 
Afterwards, $\mathcal{D}_{Train}$ is distributed among $M=100$ simulated clients, while $\mathcal{D}_{Test}$ is used solely during the subsequent evaluation of the FL models' performance. As stated earlier, we aim to simulate two different types of label imbalance, namely local and global label imbalance. Accordingly, instead of allocating samples at random, we distribute them as follows.

Our simulation of local label imbalance follows established procedures to simulate data imbalance. More precisely, our work follows Li et at. \cite{li2022federated}, who utilize a \textit{dirichlet distribution} controlled by the concentration parameter $\alpha_{local}$ to enforce label imbalance among clients. In short, the dirichlet distribution determines the proportion of samples a client holds of each class and thus controls for the severity of label imbalance among the clients, where small $\alpha_{local}$ values imply high imbalance \cite{dusing2022towards}.  In order to simulate different magnitudes of data imbalance, we decrease $\alpha_{local}$ step-wise starting with $\alpha_{local} = inf$ (perfectly homogeneous data) until $\alpha_{local} = 0.1$ (severely imbalanced data) is reached. 

Similarly, we rely on a dirichlet distribution to simulate global imbalance. Accordingly, the concentration parameter $\alpha_{global}$ controls the degree of global label imbalance. Here, the dirichlet distribution is used to determine the portion of samples per class dropped from $\mathcal{D}_{Train}$ prior to distributing them among clients. Consequently, the previously balanced combined data of the federation becomes more imbalanced with respect to the amount of samples held per class. Per default, we set $\alpha_{local} = 2$ and $\alpha_{global} = 2$ during all experiment.
\\
\textit{\textbf{Model Setting.}} For the tabular dataset CovType, we deploy an MLP consisting of three linear layers. Its input layer is of size 54, the hidden layers have a size of 45, 30 and 15, respectively, and the output layer has two neurons. Further, we apply ReLU-activation for all but the output layer and apply dropout of $0.2$. 
For \textit{MNIST}, we utilize a CNN consisting of two convolutional layers, each with a kernel size of 5. After applying 2x2 max pooling, the network stacks two linear layers (again with ReLU-activation and dropout), where the output size of the final layer is ten.
Both model architectures are inspired by Li et al. \cite{li2022federated}, who successfully applied them to various FL settings in order to empirically investigate the effects of data imbalance on FL performance. 
\\
\textit{\textbf{FL Setting.}} As outlined in Section \ref{methodology}, we apply our DC client selection to the three FL aggregation strategies \textit{FedAvg} \cite{mcmahan2017communication}, \textit{FedAtt} \cite{ji2019learning}, and \textit{FedProx} \cite{li2020federated}.
Each strategy -- regardless of whether DC is in place or not -- is trained for 100 rounds of federated training. During each round, $m=10$ active clients are randomly selected. The number of clients added through client selection is per default set to $m_{DC}=5$. All active clients perform three epochs of local training per round.
\\
\textit{\textbf{Evaluation Setting.}} To evaluate the performance of each model, we rely on the weighted F1-score that measures the F1-score per classes and weights them by their respective support.  
The reported results are mean F1-score (and standard deviation) obtained from three consecutive rounds of training and evaluation. 

\subsection{Local Imbalance}

\begin{table}[t]
\caption{Performance on local imbalance (improvements over respective baselines are underlined, best performance per dataset and $\alpha_{local}$ is bold)}
\label{tab:local_performance}
\centering
\scalebox{0.8}{
\begin{tabular}{c | c | c || c c c c c c c}
    \multicolumn{3}{c||}{} & \multicolumn{7}{c}{ $\alpha_{local}$} \\ \hline
    Data & Target & Approach & inf & 5 & 2 & 1 & 0.5 & 0.2 & 0.1 \\ 
    \hline \hline
     \parbox[t]{2mm}{\multirow{9}{*}{\rotatebox[origin=c]{90}{MNIST}}} & \parbox[t]{2mm}{\multirow{3}{*}{\rotatebox[origin=c]{90}{None}}} & $FedAvg$ &0.9511&0.9418&0.9334&0.9367&0.9193&0.8416&0.5009\\ 
     & & $FedAtt$ & 0.9402&0.9343&0.9339&0.9259&0.8858&0.8053&0.2082 \\ 
     & & $FedProx$ & 0.9500&0.9433&0.9362&0.9395&0.9324&0.8816&0.7879\\ \cline{2-10}
     
     & \parbox[t]{2mm}{\multirow{3}{*}{\rotatebox[origin=c]{90}{Real}}} & $FedAvg_{DC}$ & \underline{0.9534}&0.9244&\textbf{\underline{0.9483}}&\underline{0.9372}&\underline{0.9278}&\underline{0.8743}&\underline{0.7122}\\ 
     & & $FedAtt_{DC}$ &\underline{0.9481}&\underline{0.9401}&\underline{0.9456}&0.9097&\underline{0.9122}&\underline{0.8372}&\underline{0.4669}\\ 
     & & $FedProx_{DC}$ & 0.9490&\underline{0.9473}&\underline{0.9387}&\underline{0.9459}&\textbf{\underline{0.9410}}&\underline{0.9171}&\underline{0.8043}\\ \cline{2-10}
    
     & \parbox[t]{2mm}{\multirow{3}{*}{\rotatebox[origin=c]{90}{\fontsize{6}{10}\selectfont Balanced}}} & $FedAvg_{DC}$ & \textbf{\underline{0.9535}}&0.9399&\underline{0.9462}&\underline{0.9457}&\underline{0.9315}&\underline{0.9045}&\underline{0.7775} \\ 
     & & $FedAtt_{DC}$ & 0.9344&\underline{0.9466}&\underline{0.9431}&\underline{0.9383}&\underline{0.9346}&0.7870&\underline{0.4768} \\ 
     & & $FedProx_{DC}$ & \underline{0.9503}&\textbf{\underline{0.9498}}&\underline{0.9445}&\textbf{\underline{0.9478}}&\underline{0.9382}&\textbf{\underline{0.9201}}&\textbf{\underline{0.8048}} \\ \hline\hline

     \parbox[t]{2mm}{\multirow{9}{*}{\rotatebox[origin=c]{90}{CovType}}} & \parbox[t]{2mm}{\multirow{3}{*}{\rotatebox[origin=c]{90}{None}}} & $FedAvg$ & 0.7467&0.7318&0.7142&0.6999&0.6899&0.6649&0.5607\\
     & & $FedAtt$ & 0.7550&0.7493&0.7278&0.7176&0.6800&0.6704&0.5755\\ 
     & & $FedProx$ & 0.7495&0.7414&0.7316&0.7110&0.7023&0.6727&0.6069\\ \cline{2-10}
     
     & \parbox[t]{2mm}{\multirow{3}{*}{\rotatebox[origin=c]{90}{Real}}} & $FedAvg_{DC}$ & 0.7389&\underline{0.7326}&\underline{0.7375}&\underline{0.7310}&0.6882&\underline{0.6766}&\underline{0.6219}\\ 
     & & $FedAtt_{DC}$ & \textbf{\underline{0.7557}}&\underline{0.7542}&\underline{0.7386}&\underline{0.7314}&\underline{0.7143}&\underline{0.6928}&\underline{0.6104}\\ 
     & & $FedProx_{DC}$ & 
     \underline{0.7501}&\underline{0.7450}&\underline{0.7473}&\underline{0.7371}&\underline{0.7182}&\underline{0.6938}&\underline{0.6435}\\ \cline{2-10}
     
     & \parbox[t]{2mm}{\multirow{3}{*}{\rotatebox[origin=c]{90}{\fontsize{6}{10}\selectfont Balanced}}} & $FedAvg_{DC}$ & 0.7367&\underline{0.7338}&\underline{0.7351}&\underline{0.7363}&\underline{0.7074}&\underline{0.6918}&\underline{0.6446} \\ 
     & & $FedAtt_{DC}$ & \textbf{\underline{0.7557}}&\textbf{\underline{0.7576}}&\underline{0.7435}&\underline{0.7356}&\textbf{\underline{0.7207}}&\underline{0.7057}&\underline{0.6298} \\ 
     & & $FedProx_{DC}$ & \textbf{\underline{0.7537}}&\underline{0.7478}&\textbf{\underline{0.7495}}&\textbf{\underline{0.7426}}&\underline{0.7179}&\textbf{\underline{0.7148}}&\textbf{\underline{0.6643}} \\
\end{tabular}
}
\end{table}
To assess the potential of our proposed extension to mitigate the detrimental effect of local imbalance on the model performance, we measure the performance of the three FL strategies depending on the datasets, target distribution, and degree of imbalance (indicated by $\alpha_{local}$). Table \ref{tab:local_performance} list the F1-scores for each of the settings. The key findings from the table are the following: (1) DC client selection yields significant improvements over the respective baselines in the vast majority of settings (indicated by underlined scores), (2) the margin of improvement increases for smaller $\alpha_{local}$, indicating that our client selection is particularly beneficial when imbalance is moderate or high, (3) among the three approaches, FedProx and $FedProx_{DC}$ provide the best performance in most settings, and (4) the \textit{Balanced} target distribution yields the best performance in most conditions. These findings indicate that setting the target to \textit{Balanced} is to be preferred when facing locally imbalanced data.

\begin{figure*}[t]
    \centering
    \raisebox{43pt}{\parbox[t]{.02\textwidth}{\rotatebox[origin=c]{90}{\textbf{MNIST}}}}
    \subfloat[][Real]{\includegraphics[width=.35\textwidth]{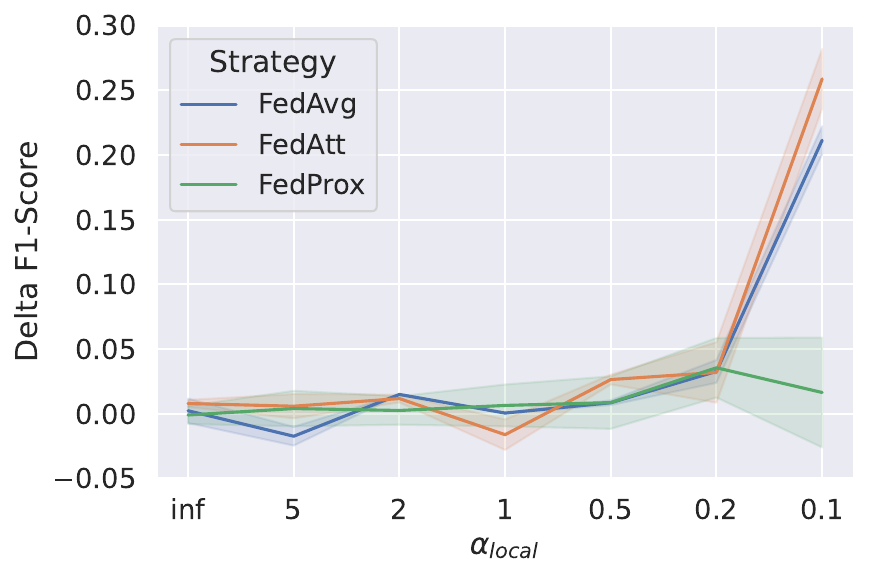}}
    \subfloat[][Balanced]{\includegraphics[width=.35\textwidth]{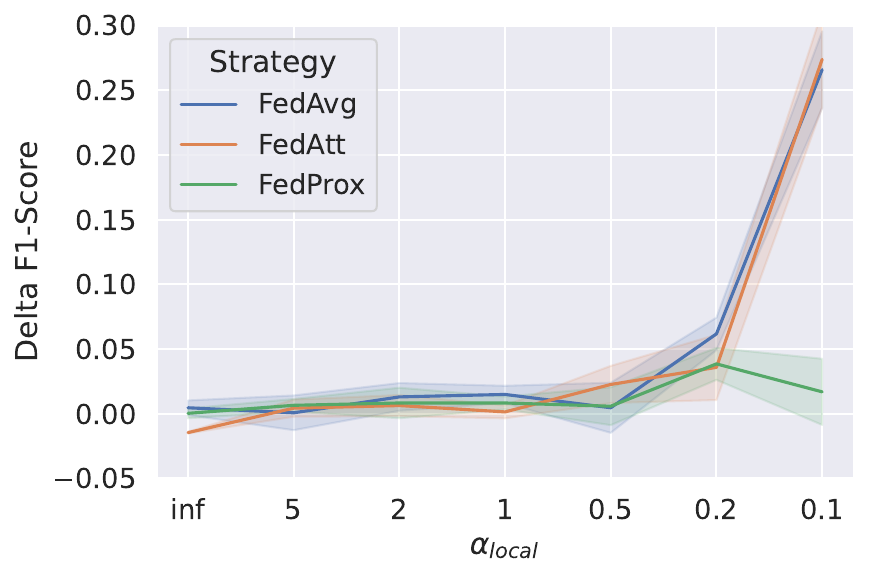}}\par
    \raisebox{43pt}{\parbox[t]{.02\textwidth}{\rotatebox[origin=c]{90}{\textbf{CovType}}}}
    \subfloat[][Real]{\includegraphics[width=.35\textwidth]{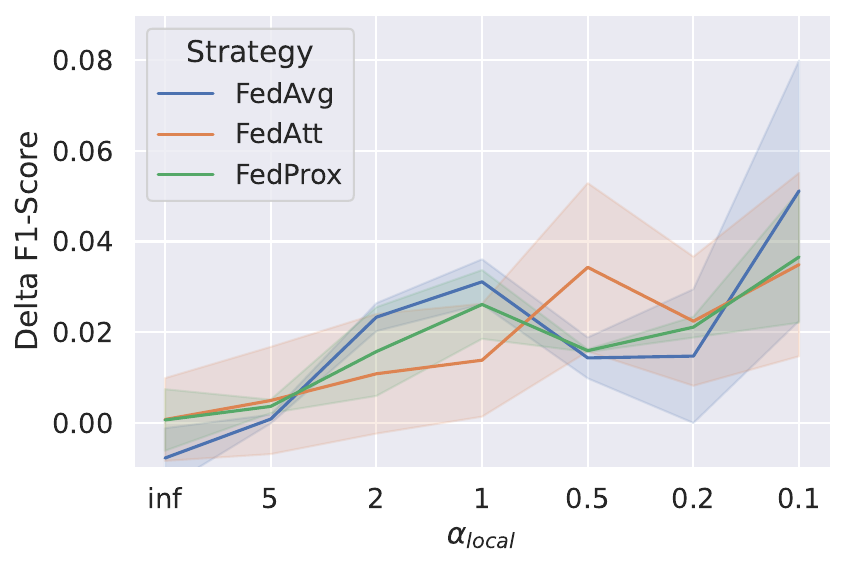}}
    \subfloat[][Balanced]{\includegraphics[width=.35\textwidth]{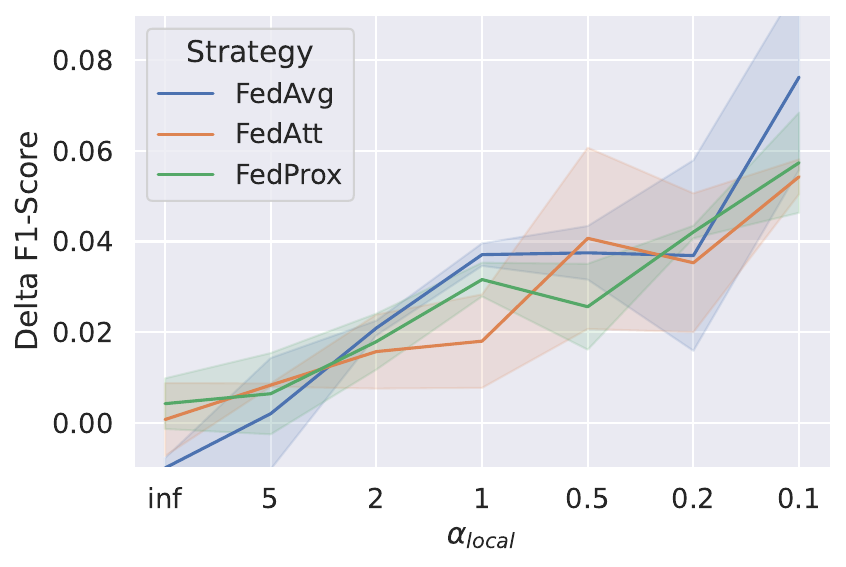}}\par
    \caption{Improvements for local imbalance $\alpha_{local}$ through distribution control}
    \label{fig:local_performance}
\end{figure*}

Figure \ref{fig:local_performance} depicts the improvements through our client selection for both datasets and target distributions. It confirms the previous finding that selecting clients to align with a \textit{balanced} distribution yields greater improvements, especially when imbalance is high ($\alpha_{local} < 0.5$). Furthermore, it shows a significant increase in F1-score improvement for a severely imbalanced FL setting ($\alpha_{local} = 0.1$). Finally, it is worth mentioning that, while all FL strategies benefit similarly from adding client selection on the CovType dataset, FedProx benefits significantly less than FedAvg and FedAtt on MNIST with $\alpha_{local} = 0.1$. This is likely due to the fact that its baseline without client selection significantly outperforms the other two strategies, leaving less room for improvements through the addition of client selection.

\subsection{Global Imbalance}

\begin{table}[t]
\caption{Performance on global imbalance (improvements over respective baselines are underlined, best performance per dataset and $\alpha_{global}$ is bold)}
\label{tab:global_performance}
\centering
\scalebox{0.8}{
\begin{tabular}{c | c | c || c c c c c c c}
    \multicolumn{3}{c||}{} & \multicolumn{7}{c}{ $\alpha_{global}$} \\ \hline
    Data & Target & Approach & inf & 5 & 2 & 1 & 0.5 & 0.2 & 0.1 \\ 
    \hline \hline
    \parbox[t]{2mm}{\multirow{9}{*}{\rotatebox[origin=c]{90}{MNIST}}} & \parbox[t]{2mm}{\multirow{3}{*}{\rotatebox[origin=c]{90}{None}}} & $FedAvg$ & 0.9441&0.9363&0.9279&0.9241&0.8896&0.7696&0.7674\\
     & & $FedAtt$ & 0.9430&0.9311&0.9306&0.9240&0.8808&0.8055&0.8056\\
     & & $FedProx$ & 0.9430&0.9378&0.9323&0.9274&0.8945&0.8364&0.8256\\ \cline{2-10}
     
     & \parbox[t]{2mm}{\multirow{3}{*}{\rotatebox[origin=c]{90}{Real}}} & $FedAvg_{DC}$ & \textbf{\underline{0.9460}}&\underline{0.9383}&\textbf{\underline{0.9483}}&\underline{0.9439}&\underline{0.8925}&\underline{0.8100}&\underline{0.8126}\\
     & & $FedAtt_{DC}$ & 0.9405&\underline{0.9439}&\underline{0.9435}&\textbf{\underline{0.9448}}&\underline{0.9073}&\underline{0.8302}&\underline{0.8061} \\ 
     & & $FedProx_{DC}$ & \underline{0.9454}&\textbf{\underline{0.9534}}&\underline{0.9450}&\underline{0.9436}&\textbf{\underline{0.9201}}&\textbf{\underline{0.8601}}&\textbf{\underline{0.8398}} \\ \cline{2-10}
     
     & \parbox[t]{2mm}{\multirow{3}{*}{\rotatebox[origin=c]{90}{\fontsize{6}{10}\selectfont Balanced}}} & $FedAvg_{DC}$ & \underline{0.9453}&\underline{0.9457}&\underline{0.9462}&\underline{0.9384}&\underline{0.8453}&\underline{0.7971}&\underline{0.7993}\\ 
     & & $FedAtt_{DC}$ & \underline{0.9448}&\underline{0.9464}&\underline{0.9468}&\underline{0.9331}&\underline{0.8906}&\underline{0.8118}&\underline{0.7961}\\ 
     & & $FedProx_{DC}$ & \underline{0.9423}&\underline{0.9408}&\underline{0.9479}&\underline{0.9312}&\underline{0.9071}&\underline{0.8414}&\underline{0.8187}\\ \hline\hline

     \parbox[t]{2mm}{\multirow{9}{*}{\rotatebox[origin=c]{90}{CovType}}} & \parbox[t]{2mm}{\multirow{3}{*}{\rotatebox[origin=c]{90}{None}}} & $FedAvg$ & 0.7430&0.7218&0.7162&0.7249&0.7064&0.6858&0.6569\\ 
     & & $FedAtt$ & 0.7399&0.7318&0.7249&0.7199&0.7254&0.7075&0.6975\\ 
     & & $FedProx$ & 0.7520&0.7482&0.7316&0.7325&0.7227&0.7016&0.6857\\ \cline{2-10}
     
     & \parbox[t]{2mm}{\multirow{3}{*}{\rotatebox[origin=c]{90}{Real}}} & $FedAvg_{DC}$ & 0.7387&\underline{0.7278}&\underline{0.7276}&0.7244&\underline{0.7285}&\underline{0.7261}&\underline{0.7237} \\
     & & $FedAtt_{DC}$ & \underline{0.7441}&\underline{0.7422}&\underline{0.7337}&\underline{0.7312}&\underline{0.7300}&\underline{0.7330}&\textbf{\underline{0.7293}} \\
     & & $FedProx_{DC}$ & \textbf{\underline{0.7602}}&\textbf{\underline{0.7524}}&\textbf{\underline{0.7494}}&\textbf{\underline{0.7521}}&\underline{0.7444}&\textbf{\underline{0.7251}}&\underline{0.7193} \\ \cline{2-10}
     
     & \parbox[t]{2mm}{\multirow{3}{*}{\rotatebox[origin=c]{90}{\fontsize{6}{10}\selectfont Balanced}}} & $FedAvg_{DC}$ & 0.7367&\underline{0.7298}&\underline{0.7271}&0.7163&0.6874&\underline{0.7006}&\underline{0.6603}\\
     & & $FedAtt_{DC}$ & 0.7386&\underline{0.7362}&\underline{0.7304}&\underline{0.7285}&0.7250&\underline{0.7204}&\underline{0.7051}\\
     & & $FedProx_{DC}$ & 0.7516&\underline{0.7489}&\underline{0.7465}&\underline{0.7488}&\textbf{\underline{0.7482}}&\underline{0.7158}&\underline{0.7023}\\
\end{tabular}
}
\end{table}
Similar to our experiments on local imbalance, we control for the impact of global imbalance on the performance of our client selection approach. Table \ref{tab:global_performance} contains the mean F1-scores for all datasets, target distributions, FL strategies, and $\alpha_{global}$ configurations.
It shows that the overall effects of global imbalance on the performance are less pronounced, as the performance decreases less compared to our previous findings on local imbalance. However, the results confirm the trend that higher imbalance (characterized by lower $\alpha_{global}$) causes weaker performance. 
Furthermore, the findings show that FedProx and $FedProx_{DC}$ are again the approaches with the best performance in most of the settings.

In contrast to our results for local imbalance, we find that setting the target distribution to \textit{Real} yields the overall best results in face of global imbalance (indicated by the vast majority of bold values in the respective rows). Using the federation's combined label distribution as target distribution is particularly beneficial when imbalance is high ($\alpha_{global} < 1$).

\begin{figure*}[t]
    \centering
    \raisebox{43pt}{\parbox[t]{.02\textwidth}{\rotatebox[origin=c]{90}{\textbf{MNIST}}}}
    \subfloat[][Real]{\includegraphics[width=.35\textwidth]{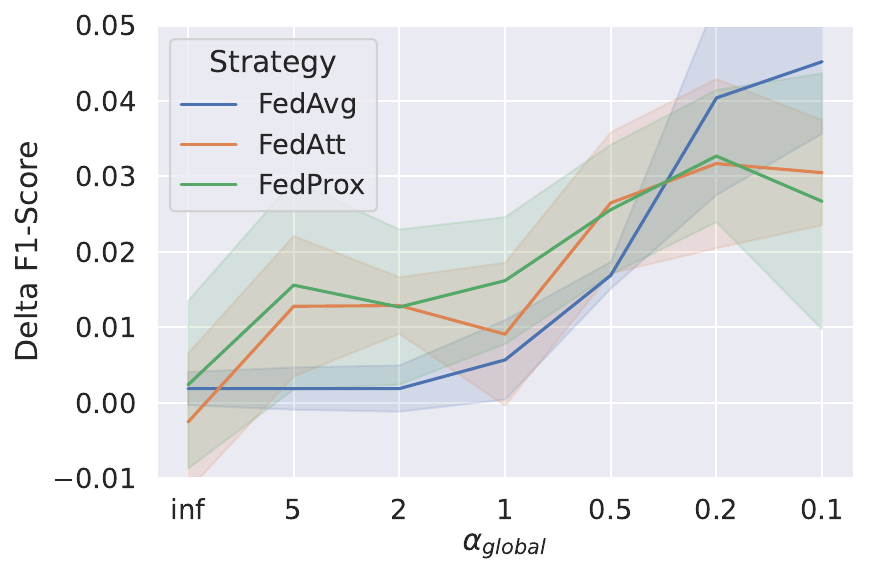}}
    \subfloat[][Balanced]{\includegraphics[width=.35\textwidth]{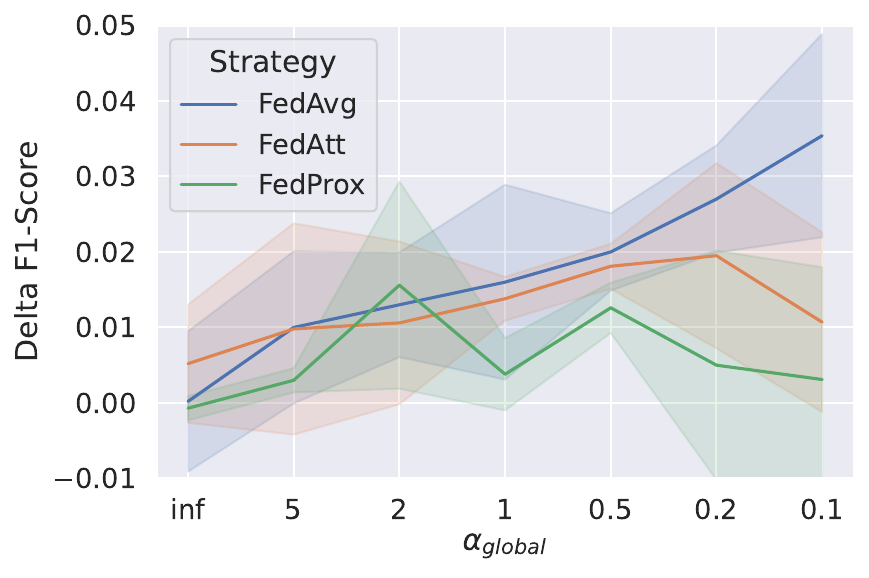}}\par
    \raisebox{43pt}{\parbox[t]{.02\textwidth}{\rotatebox[origin=c]{90}{\textbf{CovType}}}}
    \subfloat[][Real]{\includegraphics[width=.35\textwidth]{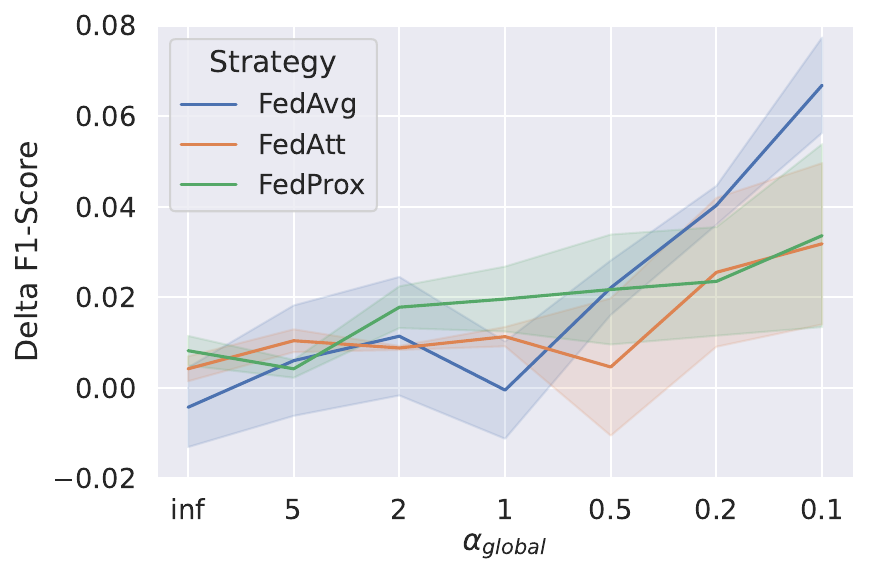}}
    \subfloat[][Balanced]{\includegraphics[width=.35\textwidth]{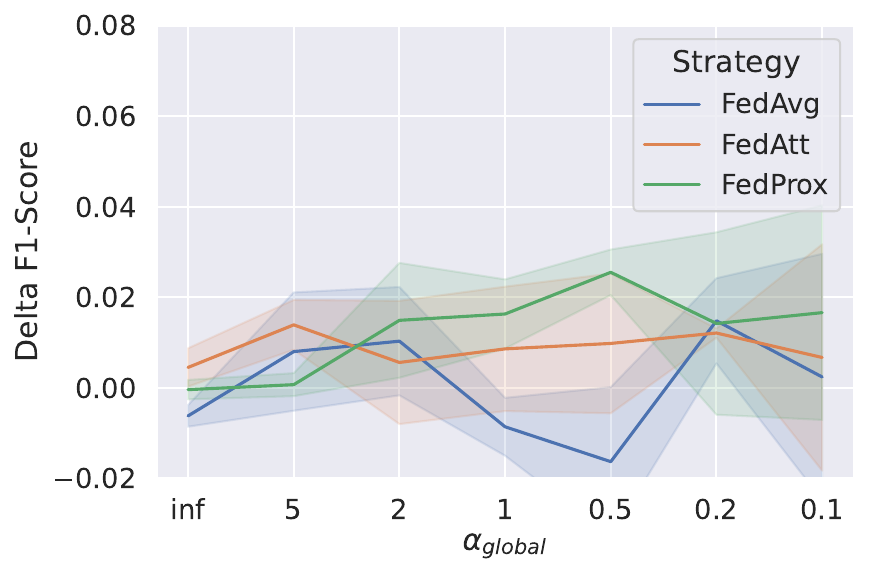}}\par
    \caption{Improvements for global imbalance $\alpha_{global}$ through distribution control}
    \label{fig:global_performance}
\end{figure*}

Figure \ref{fig:global_performance} shows the improvements through DC client selection depending on the choice of target distribution and the respective dataset. From the figure, three key observations can be made: (1) the figure confirms previous findings that the \textit{real} target distribution yields greater improvements on globally imbalanced data, (2) higher imbalance comes along with greater improvements through client selection, making it particularly suitable for FL on highly imbalanced data, and (3) unlike previously, while all FL strategies benefit from applying our client selection with \textit{real} target distribution, FedAvg shows the greatest improvements. 

\subsection{Additional Analyses}

\begin{figure*}[t]
    \centering
    \subfloat[][MNIST]{\includegraphics[width=.35\textwidth]{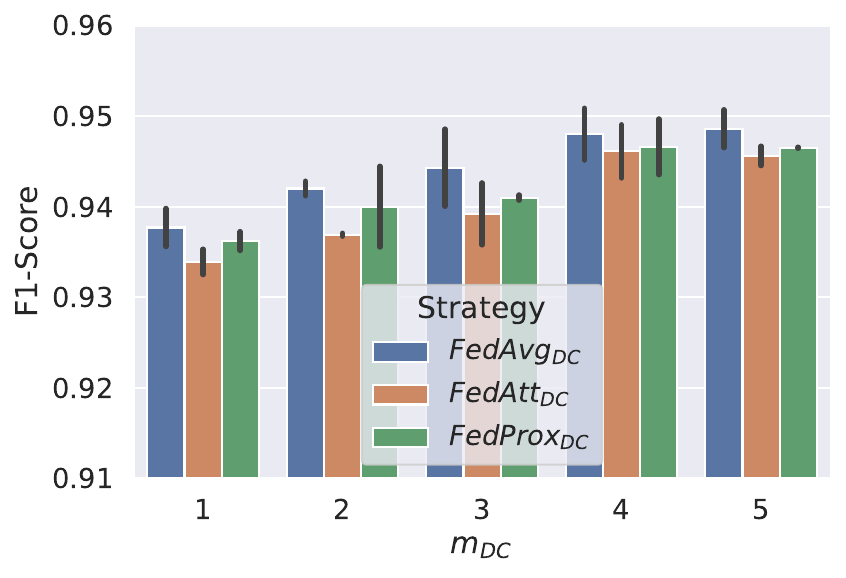}}
    \subfloat[][CovType]{\includegraphics[width=.35\textwidth]{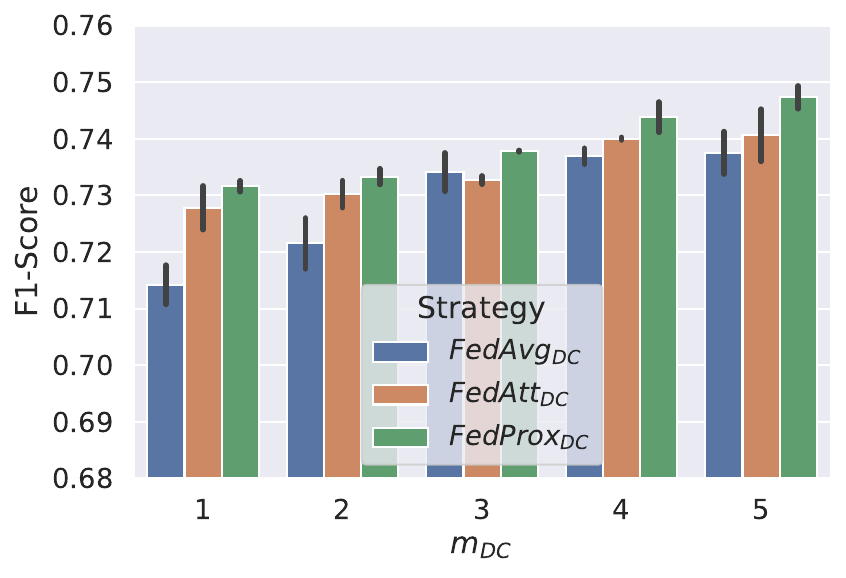}}
    \caption{Number of added clients $m_{DC}$}
    \label{fig:combinations}
\end{figure*}

This section complements our previous analyses by examining additional aspects of the proposed client selection.
For all experiments, we set $\alpha_{global} = 2$ and $\alpha_{local} = 2$ to introduce a moderate degree of data imbalance.
\\
\textit{\textbf{Number of added clients.}} We start by investigating the impact of the hyperparameter $m_{DC}$ on the performance. It determines the maximum number of clients added during each round of training. Figure \ref{fig:combinations} depicts the performance of all three FL strategies with DC client selection in place as a function of the number of added clients $m_{DC}$. Note that we chose to not include the results for both target distributions, as we found them to be very similar for all settings. Instead, we selected the better performing target distribution during all experiments.

The results show that increasing $m_{DC}$ improves the performance of $FedAvg_{DC}$, $FedAtt_{DC}$, and $FedProx_{DC}$ on both datasets. For MNIST we see a slight improvement in performance for every added client until $m_{DC} = 4$. Adding further clients does not result in significant performance improvements. In line with previous findings, we find that FedAvg benefits the most from applying client selection to it. Our results on CovType in Figure \ref{fig:combinations} (b) confirm these findings, but also show even greater improvements in terms of F1-score for each client added. Here, it is also the case that the improvement in performance stagnates from $m_{DC} = 4$ onward. Overall, these findings justify our previous choice to set $m_{DC} = 5$ for the respective data settings in this study.
\\
\textit{\textbf{Greedy and exhaustive client selection.}} In Section \ref{methodology} we outlined the greedy nature of our proposed approach that significantly reduces its computational overhead during client selection, especially in large federations. However, although it may result in a locally optimal selection of client, it may fail to find the global optimum. This is due to the fact that choosing one client after another may excludes some client combinations that would have enabled an even better approximation to the target distribution in retrospect. Therefore, we decide to compare our greedy selection approach with an exhaustive gold standard that considers all possible combinations of available clients to determine the set of clients that yield the best label distribution. 

\begin{figure*}[t]
    \centering
    \subfloat[][MNIST]{\includegraphics[width=.35\textwidth]{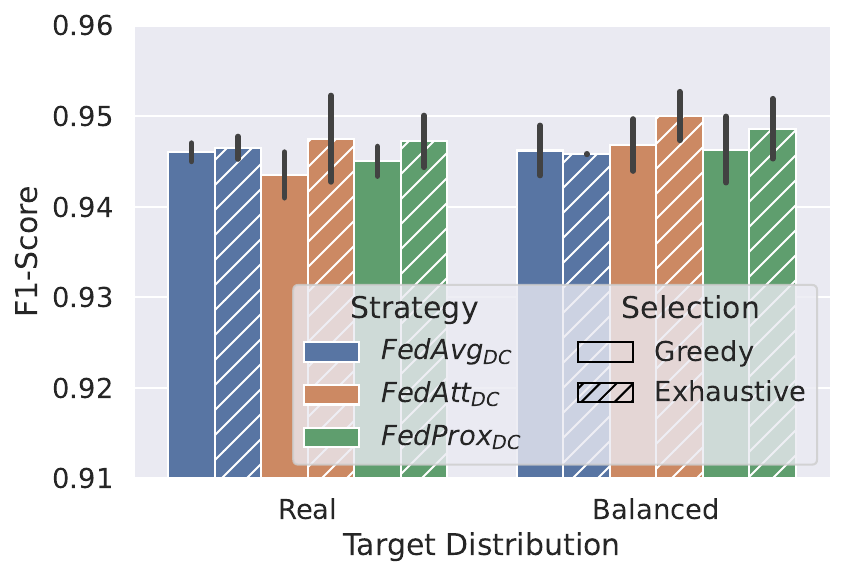}}
    \subfloat[][CovType]{\includegraphics[width=.35\textwidth]{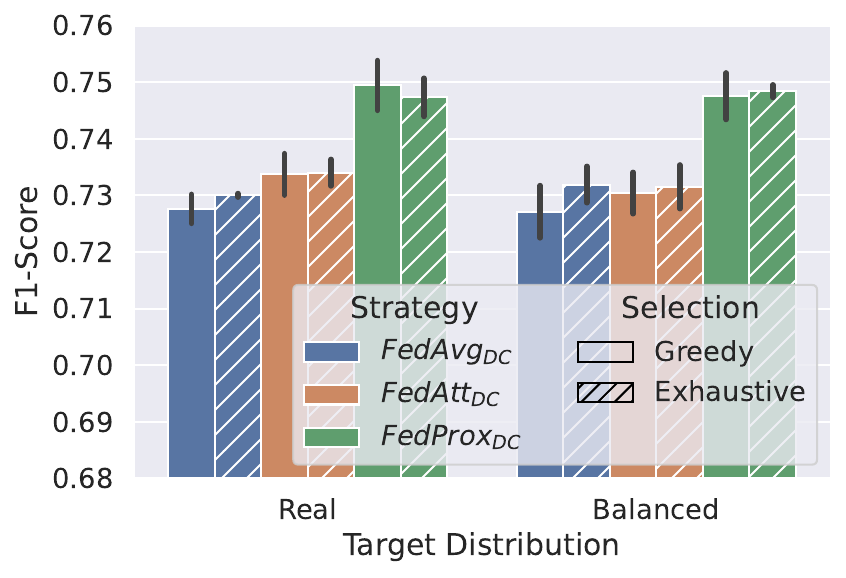}}
    \caption{Greedy and exhaustive client selection}
    \label{fig:complete}
\end{figure*}
\begin{figure*}[t]
    \centering
    \subfloat[][MNIST]{\includegraphics[width=.35\textwidth]{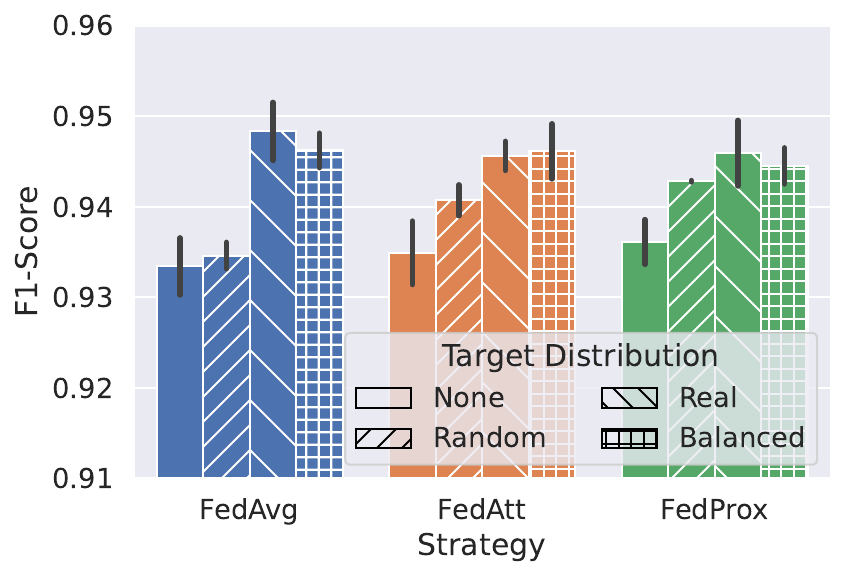}}
    \subfloat[][CovType]{\includegraphics[width=.35\textwidth]{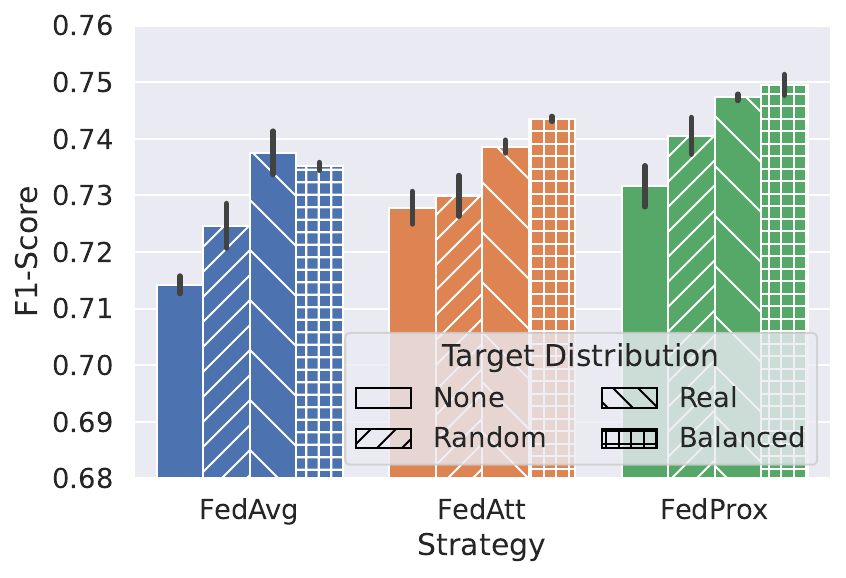}}
    \caption{Ablation study with randomized clients}
    \label{fig:ablation}
\end{figure*}

The results of this comparison are presented in Figure \ref{fig:complete}. It shows that the exhaustive client selection approach achieves marginal improvements over most of the greedy baselines on both datasets. It is worth mentioning however that most of these improvements are within their standard deviation and thus not statistically significant. Interestingly, $FedAtt_{DC}$ benefits the most from applying an exhaustive client selection compared to the other strategies. From these findings we conclude that our initial design decision to utilize greedy client selection is a feasible and efficient strategy.
\\
\textit{\textbf{Ablation study with random client selection.}} Finally, we want to make sure that the improvements through DC client selection can indeed be attributed to the proposed selection procedure and not just to the increased number of participating clients per round. Therefore, we perform an ablation study where instead of choosing the clients that provide the best alignment with the target distribution, we randomly select $m_{DC}=5$ additional clients. Thus, we match the number of active clients in $FedAvg_{DC}$, $FedAtt_{DC}$, and $FedProx_{DC}$. 

The results depicted in Figure \ref{fig:ablation} show some improvements through the addition of randomly selected clients over the respective baselines. However, among both datasets and all FL strategies, these improvements fall short of those achieved using either of the two proposed target distributions.
Among the three strategies, FedAvg benefits the most from non-random client selection. 

\section{Conclusion, Limitations, and Future Work} \label{conclusion}

In this paper, we introduce a novel DC client selection extension to existing FL strategies. Instead of choosing active clients at random, we propose to include additional clients selected using their respective label distributions. In particular, we select those clients that align the combined label distribution of all active clients best with one of two target distributions. For this purpose, we either use a \textit{balanced} label distribution or the \textit{real} label distribution of the entire federation.

Our results show that DC client selection improves the performance of all three FL strategies in face of data imbalance. In this regard, we show that the improvements increase with increasing data imbalance, making DC client selection particularly suitable for highly imbalanced FL environments. 
Furthermore, we uncover distinctive advantages of using either target distribution, where the choice of a \textit{balanced} distribution offers larger improvements amidst local imbalance, whereas the \textit{real} distribution is more beneficial facing global imbalance.
Our additional analyses prove that (1) while some of the improvements through our method can be attributed to the increased number of active clients per round, DC client selection is responsible for most of the improvements, (2) the number of added clients affects the margin of improvements, and (3) our greedy client selection approach performs similarly to the exhaustive gold-standard, which justifies our initial choice for a greedy approach given its superior scalability.
\\
\textit{\textbf{Limitations.}} Currently, the presented method requires all clients to truthfully announce their cosine distance score during client selection. Accordingly, malicious clients could compromise the federated training procedure by untruthfully reporting minimal distances, which results in their frequent selection. 

Furthermore, our current evaluation is limited to local and global label imbalance. However, additional nuances of data imbalance, such as feature imbalance and quantity imbalance have detrimental effects on FL, too \cite{dusing2022trade,wang2021addressing}. Moreover, artificially splitting data among simulated clients might fall short of capturing the complex nature of data imbalance in real-world federated datasets \cite{ogier2022flamby}.
\\
\textit{\textbf{Future Work.}} In an effort to make our proposed DC client selection more robust to malicious or faulty clients, we plan to utilize various methods from the field of federated analytics \cite{wang2021federated} to replace the error-prone cosine distance self-reporting step with a privacy-preserving procedure issued by the central server.

Moreover, some follow-up work will utilize our findings to develop a self-adjusting client selection approach that is capable of dynamically adjusting its target distribution to the data present at the respective federation.

\end{document}